\documentclass[orivec]{llncs}

\usepackage{eccv}

\usepackage{eccvabbrv}

\usepackage{graphicx}
\usepackage{booktabs}

\usepackage{wrapfig}
\usepackage{tikz}

\usepackage[accsupp]{axessibility}  % Improves PDF readability for those with disabilities.

\usepackage[pagebackref,breaklinks,colorlinks,citecolor=cyan]{hyperref}
\usepackage{orcidlink}

\usepackage{epsfig}
\usepackage{graphicx}
\usepackage{tikz}
\usepackage{pgfplots}
    \pgfplotsset{compat=1.18}
\usepackage{caption}
\usepgfplotslibrary{groupplots}
\usetikzlibrary{patterns}

\usepackage{fdsymbol}

\usepackage{lipsum}

\usepackage{graphicx}
\usepackage{amsmath}
\usepackage{amssymb}
\usepackage{booktabs}

\usepackage{arydshln}

\usepackage{tikz}
\usepackage{comment}
\usepackage{amsmath,amssymb} % define this before the line numbering.
\usepackage{color}
\usepackage{multirow}
\usepackage{makecell}
\usepackage{bbding}
\usepackage[colorlinks]{hyperref}
\usepackage{color, colortbl}

\usepackage{algpseudocode}
\usepackage{algorithm}

\usetikzlibrary{arrows}
\usepackage{tkz-kiviat}
\usepackage{xfp}

\usepackage{pgfkeys}
	\def\addlegendimage{\csname pgfplots@addlegendimage\endcsname}

\usepackage{arydshln}

\usepackage[capitalize]{cleveref}
\crefname{section}{Sec.}{Secs.}
\Crefname{section}{Section}{Sections}
\Crefname{table}{Table}{Tables}
\crefname{table}{Tab.}{Tabs.}

\definecolor{citecolor}{HTML}{0071bc}
\definecolor{color_ao}{gray}{0.5}
\definecolor{color_our}{rgb}{0.66,0.82,0.56}
\definecolor{color_pre}{rgb}{0.52,0.59,0.69}
\definecolor{Gray}{gray}{0.9}
\definecolor{LighterGray}{gray}{0.93}
\definecolor{LightGrayForTableRule}{gray}{0.92}
\definecolor{DarkGray}{gray}{0.5}
\definecolor{Black}{rgb}{0.0, 0.0, 0.0}
\definecolor{NiceBlue}{rgb}{0.11764705882352941, 0.5647058823529412, 1.0}

\definecolor{demphcolor}{RGB}{114,114,114}
\newcommand{\demph}[1]{\textcolor{demphcolor}{#1}}
\definecolor{E6F5F0}{HTML}{f7fcf9}
\definecolor{6ec2b5}{HTML}{6ec2b5}
\definecolor{FFABA8}{HTML}{fc9d9a}
\definecolor{darkF7E0D5}{RGB}{209,154,128}
\colorlet{Light}{E6F5F0}
\colorlet{Dark}{6ec2b5}
\colorlet{Salmon}{FFABA8}

\definecolor{e9f7f2}{HTML}{f7fcf9}
\colorlet{LightColor}{e9f7f2}

\newcommand{\CC}[1]{\cellcolor{Light}}

\usepackage{amssymb}% http://ctan.org/pkg/amssymb
\usepackage{pifont}% http://ctan.org/pkg/pifont

\usepackage{tcolorbox}
\newtcolorbox{abox}{colback=LightColor,colframe=Dark}

\newcommand{\modelname}{\textsc{SafaRi}}
\newcommand{\MVF}{\textsc{SpARC}}
\newcommand{\FFE}{\textsc{X-FACt}}

\begin{document}

\title{\modelname:Adaptive \underline{S}equence Tr\underline{a}ns\underline{f}ormer for We\underline{a}kly Supervised \underline{R}eferring Expression Segmentat\underline{i}on} 

\titlerunning{\modelname}

\author{Sayan Nag\inst{1,2}\thanks{Work done during internship at Adobe Research.} \and
Koustava Goswami\inst{2} \and
Srikrishna Karanam\inst{2}}

\authorrunning{S.~Nag et al.}

\institute{University of Toronto \\
\email{sayan.nag@mail.utoronto.ca}\\
\and
Adobe Research\\
\email{\{koustavag,skaranam\}@adobe.com} \\
\tt\small Project page -\ \url{https://sayannag.github.io/safari_eccv2024/}}

\maketitle

\begin{abstract}
  Referring Expression Segmentation (RES) aims to provide a segmentation mask of the target object in an image referred to by the text (i.e., referring expression). Existing methods require large-scale mask annotations. Moreover, such approaches do not generalize well to unseen/zero-shot scenarios. To address the aforementioned issues, we propose a weakly-supervised bootstrapping architecture for RES with several new algorithmic innovations. To the best of our knowledge, ours is the first approach that considers only a fraction of both mask and box annotations (shown in Figure \ref{fig:teaser} and Table \ref{tab:comparison_overview}) for training. To enable principled training of models in such low-annotation settings, improve image-text region-level alignment, and further enhance spatial localization of the target object in the image, we propose Cross-modal Fusion with Attention Consistency module. For automatic pseudo-labeling of unlabeled samples, we introduce a novel Mask Validity Filtering routine based on a spatially aware zero-shot proposal scoring approach. Extensive experiments show that with just 30\% annotations, our model \modelname\ achieves 59.31 and 48.26 mIoUs as compared to 58.93 and 48.19 mIoUs obtained by the fully-supervised SOTA method SeqTR respectively on RefCOCO+@testA and RefCOCO+testB datasets. \modelname\ also outperforms SeqTR by 11.7\% (on RefCOCO+testA) and 19.6\% (on RefCOCO+testB) in a fully-supervised setting and demonstrates strong generalization capabilities in unseen/zero-shot tasks.
  % Code will be publicly released.
  \keywords{Weakly-Supervised Referring Expression Segmentation \and Attention Consistency \and Mask Validity Filtering}
\end{abstract}

\section{Introduction}
\label{sec:intro}
We consider the problem of referring expression segmentation (RES) \cite{zhu2022seqtr,busnet,vlt,cmsa,liu2017recurrent,li2018referring,margffoy2018dynamic,polyformer,shi2018key,chen2019see,ye2019cross,hu2020bi,vistallm} where given an image and a text sentence, the goal is to segment out parts of the image the text is referring to. Naturally, much existing work in RES blends vision-language understanding \cite{lu2019vilbert,li2020oscar,lu202012,zhang2021vinvl,uniter,yang2022unitab,radford2021learning,dou2022empirical} and instance segmentation \cite{liu2018path,chen2019hybrid,dai2016instance,he2017mask,bolya2019yolact}, allowing for object segmentation using free-form textual expressions rather than predefined categories \cite{hui2020linguistic,huang2020referring,feng2021encoder,vlt,yang2022lavt,wang2022cris,kim2022restr, hu2016segmentation}. Existing RES methods are predominantly \textit{fully-supervised} in nature \cite{vlt,cmsa,busnet,huang2020referring,feng2021encoder,hu2016segmentation,vlt,shi2018key,liu2017recurrent,chen2019see,zhu2022seqtr,polyformer}. Hence, they necessarily require expensive-to-obtain and extensive manually-annotated segmentation Ground-Truth (GT) masks for training and typically do not generalize well to unseen/zero-shot scenarios.

\begin{figure}[!t]
    % \vspace{-1.0cm}
    \centering
    \includegraphics[width=\textwidth]{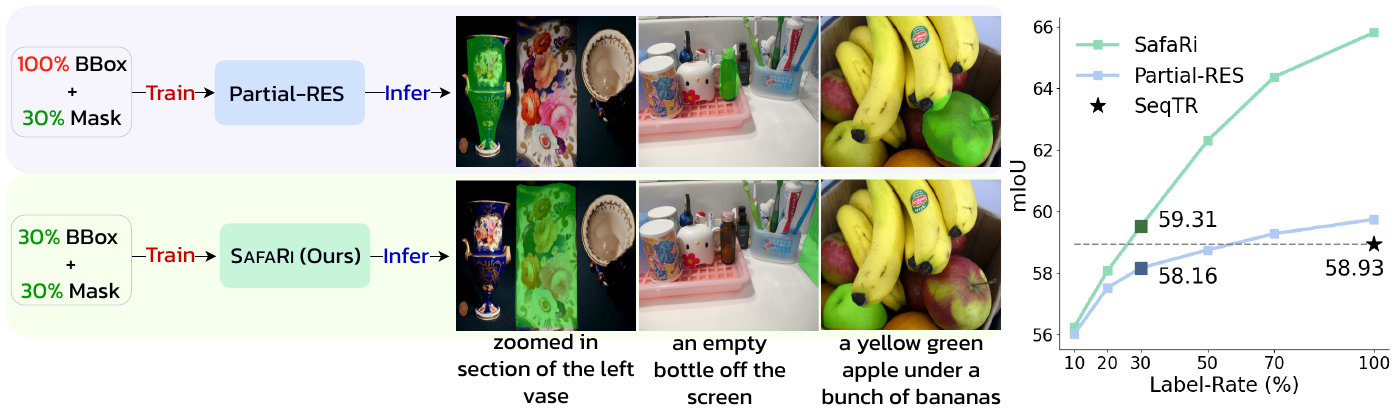}
    \caption{\textbf{\modelname\ achieves state-of-the-art performance} in both weakly- and fully-supervised RES tasks. Although unlike Partial-RES, \modelname\ is not pretrained on fully-supervised REC task, with just 30\% annotations, \modelname\ achieves 59.31 mIoU whereas Partial-RES and fully-supervised SeqTR obtains 58.16 and 58.93 mIoUs (see mIoU vs Label-Rate plot). In the weak-supervision setting, the inclusion of \FFE\ (with cross-modal fusion and AMCR components) and \MVF\ modules aids \modelname\ to demonstrate excellent grounding capabilities under challenging scenarios where Partial-RES fails (see qualitative examples). Quantitative results are provided in Tables \ref{table:sotaonres}-\ref{table:davishmdb}.}
    \label{fig:teaser}
    \vspace{-5mm}
\end{figure}

\begin{figure}[!t]
  \begin{minipage}[b]{0.51\linewidth}
    \centering    \includegraphics[width=0.9\linewidth, height=0.59\linewidth]{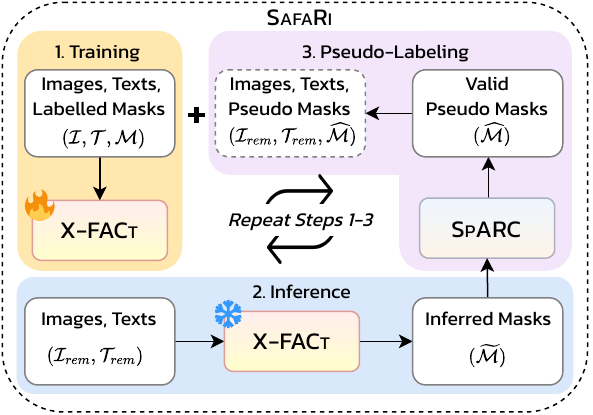}
    \captionof{figure}{\textbf{Overview of weakly-supervised bootstrapping setup}. It includes an initial training stage, followed by inference and pseudo-labeling steps. Filtered pseudo-masks are added to the initial dataset and model is retrained in an iterative manner.}
    \label{fig:bootstrap}
  \end{minipage}\hfill
\begin{minipage}[b]{0.46\linewidth}
    \centering
    % \vspace{-15mm}
    % \renewcommand{\arraystretch}{0.85}
    % \small
    \centering
    \renewcommand{\arraystretch}{1.4}
    \resizebox{\linewidth}{!}{%
    \begin{tabular}{l|cc|c|c}
    \toprule
     \multirow{2}{*}{\textbf{Method}} &
      \multicolumn{2}{c|}{\centering \textbf{100\% GT Labels}} & \textbf{Fully-supervised}
       & \textbf{Filtering with}\\
       \cmidrule{2-3}
     & \textbf{Mask?} & \textbf{BBox?} & \textbf{REC Pretraining?}
      & \textbf{100\% GT BBox?} \\ \midrule
    % \rowcolor{TableAlternateColour}
    Partial-RES \cite{qu2023learning} &
      \textcolor{OrangeRed}{\ding{55}} &
      \textcolor{ForestGreen}{\ding{51}} &
      \textcolor{ForestGreen}{\ding{51}} &
      \textcolor{ForestGreen}{\ding{51}} \\
    \rowcolor{Light} 
    \textbf{\modelname\ (Ours)} &
      \textcolor{OrangeRed}{\ding{55}} &
      \textcolor{OrangeRed}{\ding{55}} &
      \textcolor{OrangeRed}{\ding{55}} &
      \textcolor{OrangeRed}{\ding{55}} \\ \bottomrule
    \end{tabular}%
    }
    % \vspace{-2.0mm}
    \captionof{table}{\textbf{Comparison of \modelname\ with Partial-RES} \cite{qu2023learning}. Unlike Partial-RES, \modelname\ \textbf{\textit{does not use 100\% bounding boxes}} and considers \underline{mask \% $=$ box \%} and it does not involve fully-supervised REC pretraining. Furthermore, as opposed to Partial-RES, for the filtering step, \modelname\ involves a zero-shot approach to generate boxes instead of using 100\% Ground Truth (GT) boxes. All \textcolor{OrangeRed}{\ding{55}} evinces more challenging and realistic WSRES problem setting.}
    % \vspace{-8mm}
\label{tab:comparison_overview}
  \end{minipage}
  \vspace{-5mm}
\end{figure}

As noted above, obtaining large-scale GT masks is very expensive and tedious which makes it challenging to scale up such methods. To address this, a recent work (Partial-RES) \cite{qu2023learning} propose a partially (weakly) supervised solution for RES task. However, they consider abundant (100\%) bounding box annotations to begin with (Figure \ref{fig:teaser} and Table \ref{tab:comparison_overview}) and pretrain the model with these boxes on the Referring Expression Comprehension (REC) task (text referred box prediction) in a \textit{fully-supervised} manner. Moreover, they use these 100\% box annotations for filtering pseudo-labels. Their limitations are: (i) in a more pragmatic and true weakly-supervised setting \textit{the percentage of boxes should equal the percentage of masks}, (ii) in their pretraining stage the model is already aware of the grounding information of the same dataset used in the weak-supervision stage, (iii) their approach does not account for the cross-modal region-level interaction between the image and language features which is crucial for localization tasks \cite{glip,glipv2}.

On the contrary, we investigate a more realistic, challenging and unexplored problem of Weakly-Supervised Referring Expression Segmentation (WSRES) with limited human-annotated mask and box annotations, specifically where \textbf{\textit{box \% equals mask \%}} (Figure \ref{fig:teaser}, see Table \ref{tab:comparison_overview} for comparisons). To tackle our proposed novel WSRES task, we present \textbf{\modelname}, an auto-regressive contour-prediction-based RES method capable of demonstrating excellent performance under challenging scenarios with few available mask and box annotations. We specifically choose segmentation as an auto-regressive point prediction task along the object's contour instead of an independent pixel prediction approach. This is because the latter does not account for relationships between neighbouring pixels and therefore lacks structural information of the object being segmented, resulting in poor performances in the absence of abundant mask annotations \cite{qu2023learning}.

In \modelname, firstly, we introduce a novel Cross-modal (\textbf{X-}) \textbf{F}usion with \textbf{A}tte-ntion \textbf{C}onsis\textbf{t}ency (\textbf{\FFE}) module for facilitating excellent inter-domain alignment. Since we have only few annotated data, we make the cross-fusion component in \FFE\ to be flexible and parameter efficient. Further, in \FFE\ we actively leverage the cross-attention heatmaps by formulating a regularization technique that encourages these heatmaps to be consistent with the referred object in the image. This is enabled by confining the attended regions within the object contour, enhancing the fidelity of predicted masks. This is particularly beneficial for the limited annotation scenarios (in WSRES) where it explicitly helps features improving visual grounding, without relying on lots of data. Secondly, we systematically devise a novel bootstrapping strategy (Figure \ref{fig:bootstrap}) which uses a handful of labeled masks (e.g., 10\%) and iteratively trains the model by utilizing pseudo-masks obtained from a pseudo-labeling procedure. Finally, in contrast with conventional pseudo-labeling pipelines which directly use the model predictions as pseudo-labels, we design a new \textit{Mask Validity Filtering} (\textit{MVF}) routine that is responsible for selection of pseudo-masks (for unannotated data) by validating whether they spatially align with the boundaries (boxes) of the referred objects or not. Since we do not possess 100\% bounding boxes, we propose \textbf{\MVF}, a novel REC technique with spatial reasoning capabilities for obtaining these boxes in an entirely \textit{zero-shot} manner (Figure \ref{fig:mainfig}). Equipped with the above modules, unlike Partial-RES, our system learns meaningful, transferable and generalizable representations with rich semantic understanding, empowering it to reason and make accurate predictions on unseen data (Table \ref{table:davishmdb}).

We summarize our \textbf{main contributions} as: \textbf{(i)} To the best of our knowledge, ours is the first to consider an \textit{accurate representation} of Weakly-Supervised Referring Expression (WS-RES) task by considering a novel, more practical and challenging scenario with limited box and mask annotations where \underline{box \% equals} \underline{mask \%}. \textbf{(ii)} The novel \FFE\ module fosters prediction of high quality masks by improving \textit{cross-modal alignment} quality, especially where abundant ground-truth annotations are not present. \textbf{(iii)} Utilizing \MVF, a novel zero-shot REC technique, the mask validity filtering stage together with the bootstrapping pipeline improve system's \textit{self-labeling capabilities}. \textbf{(iv)} Extensive experiments demonstrate the efficacy of \modelname\ as it \textit{significantly outperforms} baseline models on RES benchmarks. \modelname\ also demonstrates \textit{strong generalization capabilities} when evaluated on an \textit{unseen} referring video object segmentation task in a \textit{zero-shot} manner. Notably, with just 30\% mask-annotated data, \modelname\ achieves 59.31 and 55.83 mIoUs versus 58.93 and 55.64 mIoUs obtained by the fully-supervised SOTA method SeqTR \cite{zhu2022seqtr} on RefCOCO+@testA and RefCOCOg@test sets (Table \ref{table:sotaonres} and Figure \ref{fig:teaser}).

% For a \textit{Supervised} REC task, typical approaches include two-stage methods employing region-proposal ranking based on textual descriptions \cite{zhang2018grounding,zhuang2018parallel,hu2017modeling}, and single-stage methods which fuse image and textual features and directly predict the bounding boxes \cite{ li2021referring,zhu2022seqtr,yang2019fast,cai2022x,zhu2022seqtr,mdetr}.
% Pseudo-Q \cite{pseudoq} produces linguistic descriptions for various objects in an image, used to train a REC model. However, such an approach can not be considered as fully unsupervised since the pseudo descriptions are obtained using a captioning generator that has been trained on the COCO dataset.
% Referring Expression Segmentation (RES) is another multi-modal grounding task which involves fine-grained pixel-level understanding of the object referred to by the text \cite{zhu2022seqtr,busnet,vlt,cmsa,liu2017recurrent,li2018referring,kim2022restr,yang2022lavt,li2021referring}.

% SeqTR \cite{zhu2022seqtr} is one such approach which unifies visual grounding tasks (REC and RES) under the same framework by conceptualizing segmentation (and bounding-box detection) as an auto-regressive contour point prediction problem. Despite its success, SeqTR uses a simple fusion strategy which does not incorporate fine-grained multi-modal alignment especially required in the presence of limited mask and bounding box annotations.

\section{Related Work}
\label{sec:related_work}
% \noindent\textbf{Referring Expression Segmentation (RES).}
\textbf{Fully-supervised RES.} These methods chiefly focus on enhancing the quality of multi-modal fusion of extracted vision-language features \cite{hu2016segmentation, liu2017recurrent}. Recent efforts have been made to develop powerful multi-modal transformer-based methods \cite{ding2021vlt, wang2022cris, li2021referring, kim2022restr,polyformer,zhu2022seqtr} showing significant improvements over previous baselines.\\
\noindent\textbf{Weakly Supervised RES.} Weakly supervised instance segmentation methods have also been investigated in the recent past \cite{biertimpel2021prior,fan2020commonality,zhou2020learning}. However, in the context of RES, there is one recent study (Partial-RES) which adopts partial or weak-supervision \cite{qu2023learning}. They specifically showed that employing a unified multi-modal transformer model, e.g., SeqTR \cite{zhu2022seqtr} is more beneficial (as compared to MDETR \cite{mdetr}) in the presence of limited annotations. In their weakly supervised RES setting, Partial-RES \cite{qu2023learning} considered abundant bounding box annotations albeit limited mask annotations. They utilized a fully-supervised REC pretrained model to train on limited mask-annotations (Table \ref{tab:comparison_overview}). Furthermore, for filtering pseudo-masks, they again used the fully available (100\%) bounding box annotations. However, since this involved 100\% box annotations, our proposed setup is much more challenging and truly weakly supervised since we consider \emph{\textbf{both} box and mask annotations to be limited.}
%We claim that this is an inaccurate setup of the Weakly-Supervised RES (WSRES) task since it involves the presence of 100\% bounding box annotations in the training data. In a true weakly-supervised setup, both the mask and box annotations should be limited (the numbers being equal).
In their pseudo-labeling step on the data without masks (but with boxes), the model has already seen the boxes in the REC pretraining stage and thus the grounding information is already present. %Such information retention effect is more pronounced, since they are using a unified model such as SeqTR \cite{zhu2022seqtr}.\\
\noindent\textbf{Unsupervised Referring Expression Comprehension (REC).} Referring Expression Comprehension (REC) is a grounding task which involves localizing an object present in an image with respect to a textual description of that object \cite{mdetr,zhuang2018parallel, hong2019learning,liao2020real,yang2019fast,volta}. With the introduction of Vision-Language Pretrained (VLP) models \cite{radford2021learning, li2021align, uniter, apollo}, it has been possible to develop \textit{Unsupervised} or \textit{Zero-Shot} REC (ZS-REC) methods \cite{cpt,reclip,redcircle,pseudoq}. CPT \cite{cpt} colors region proposal boxes and utilizes a captioning model for predicting the colored proposals that are linked to the textual expressions. RedCircle \cite{redcircle} employs visual prompting by drawing circular contours outside the detected object proposals and subsequently ranking them based on the obtained CLIP scores. However, it majorly lacks the understanding of the spatial relationships among the detected proposals.\\
\noindent\textbf{Referring Video Object Segmentation.}  Referring Video Object Segmentation (R-VOS) is a cross-modal task with the objective to segment the target object in all video frames, referred by a linguistic description. Existing approaches comprises of (i) \textit{bottom-up} methods where RES algorithms are applied independently at the frame-level \cite{cmsa, seo2020urvos}, (ii) \textit{top-down} methods where a language grounding model selects the best object tracklet from a candidate set of tracklets initially constructed by propagating the detected object masks from key frames \cite{liang2021rethinking}, and (iii) \textit{query-based} method which introduce a small set of object queries that are conditioned on the referring expression for the target object \cite{referformer}. However, most of these methods are trained in a fully-supervised manner on the benchmark R-VOS datasets. Conversely, in this work we consider a Zero Shot R-VOS (ZS-R-VOS) task where we apply our model (trained on RES task) in a \textit{zero-shot} manner.\\
\noindent\textbf{Sequence-to-Sequence (seq2seq) Modeling in Vision Tasks.} Seq2seq modeling has displayed immense success in Natural Language Processing (NLP) tasks \cite{seq2seq, seq2seq2,radford2019language,touvron2023llama,brown2020language,raffel2020exploring}. Taking inspiration from such studies, recent progress have been made to model vision tasks in a seq2seq manner \cite{pix2seq,chen2022unified,yang2022unitab, wang2022unifying,lu2022unifiedio,polyformer,zhu2022seqtr}. One of the foremost studies in this domain is Pix2Seq \cite{pix2seq} that proposes object detection as a seq2seq modeling task conditioned on the observed pixel inputs. Pix2Seqv2 \cite{chen2022unified} is an extension of Pix2Seq with an inclusion of instance segmentation and captioning tasks unified into a single shared interface. UniTAB \cite{yang2022unitab} employs a unified seq2seq learning framework which is able to jointly output open-ended text and box representations, facilitating alignment between words and boxes. Obj2Seq \cite{chen2022obj2seq} takes objects as inputs and outputs human pose estimation into sequence-generated form. Finally, SeqTR \cite{zhu2022seqtr} proposes to combine visual grounding tasks under a unified framework. Taking inspiration from these studies, we implement RES with  a contour prediction approach, however, in a weakly-supervised setting.

\section{\modelname}
\label{sec:method}

% \subsection{\modelname}
We present \modelname, an adaptive multi-modal sequence transformer with our novel components: (i) Cross-modal (\textbf{X-}) \textbf{F}usion with \textbf{A}ttention \textbf{C}onsis\textbf{t}ency (\textbf{\FFE}) module, (ii) bootstrapped Weak-Supervision with $\gamma$-Scheduling (WSGS), and (iii) Mask Validity Filtering (MVF) with \MVF.

\subsection{\FFE.} \FFE\ consists of \textit{Fused Feature Extractors} (with cross-modal fusion) and \textit{Attention Mask Consistency Regularization} (AMCR) as discussed below:\\
\noindent \textbf{3.1.1 Fused Feature Extractors.}
We use Swin transformer \cite{liu2021swin} and RoBERTa \cite{liu2019roberta} as our image and text feature extractors. A simple concatenation-based encoding strategy fails to capture cross-modal interactions, leading to poor fine-grained multi-modal representations. To address this, previous Vision-Language studies employed cross-modal fusion modules \cite{dou2022empirical,li2021align,glip} in their respective frameworks. However, such strategies introduce extra trainable fusion-specific layers. For instance, the early fusion scheme adopted in GLIP \cite{glip}, used 6 vision-language fusion layers and 6 additional BERT layers for feature enhancement. However, such a scheme is not suitable in the presence of limited annotations. Therefore, as shown in Figure \ref{fig:mainfig}, we introduce a simple and lightweight fusion mechanism through normalized gated cross-attention into the layers of uni-modal feature extractors to learn high-quality language aware visual representations:

\begin{figure*}[t]
    \centering
    \includegraphics[width=\textwidth]{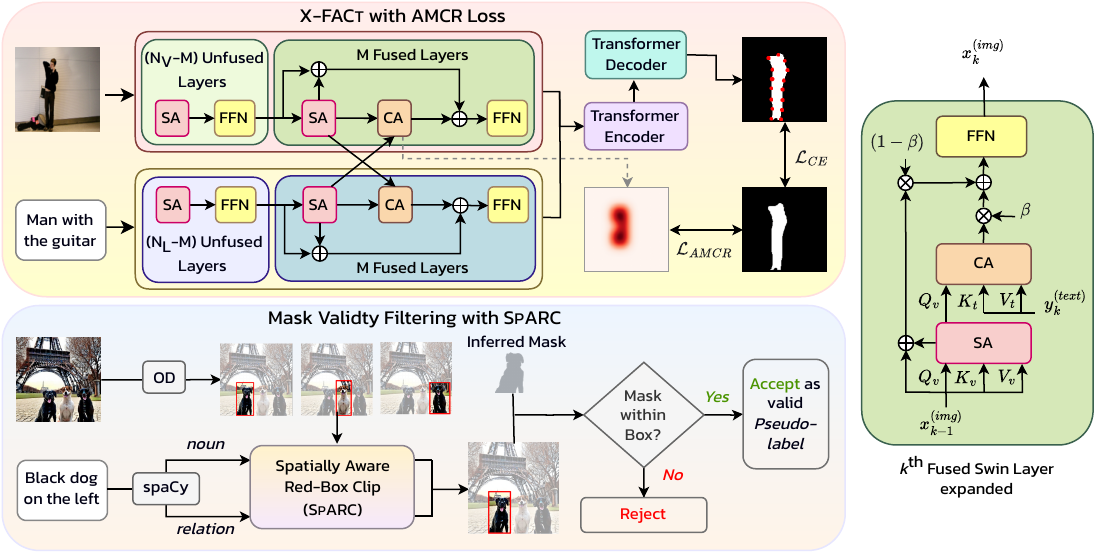}
    \caption{\textbf{Architectural components of \modelname.} (i) We introduce \FFE, composed of normalized gated cross-attention based Fused Feature Extractors and Attention Consistency Mask Regularization (AMCR) for enhancing cross-modal synergy and spatial localization of target objects. The fused output is subsequently fed to Sequence Transformer for prediction of contour points.(ii) We design Mask Validity Filtering (MVF) strategy for choosing valid pseudo-masks using \MVF\ module which is a Zero-Shot REC approach with spatial reasoning capabilities.}
    \label{fig:mainfig}
\end{figure*}

\vspace{-5mm}
\begin{gather}\label{eq:cross_attention}
    % \Tilde{x}_{k} &= \text{S-MHA}(x_{k-1}) \nonumber \\
    x_{k} = x_{k} + (1-\beta) \cdot \text{S-MHA}(x_{k-1}) + \beta \cdot \text{C-MHA}(\text{S-MHA}(x_{k-1}), y_{k})\\
    x_{k} = x_{k} + \text{FFN}(x_{k}) \nonumber
\end{gather}
where $x_{k-1}$ is the output from the $(k-1)^{th}$ layer, $\text{S-MHA}$ and $\text{C-MHA}$ represent Self- and Cross-Multi-Head Attention, FFN denotes cross-feed-forward network, and $\beta$ is a learnable weighted gating parameter initialized from 0. The fused feature then goes into a sequence transformer for decoding the object contour points as outputs. It is worth noting that unlike previous works \cite{dou2022empirical,li2021align} our gated methodology preserves the uni-modal embeddings (i.e., self-attention features), ensuring to learn the weighted representations during training. It also ensures to map linguistic semantic features to the localized parts of the images, making the system capable of understanding fine-grained visual features of the objects.
\begin{figure}[!t]
    \centering
    \includegraphics[width=\textwidth]{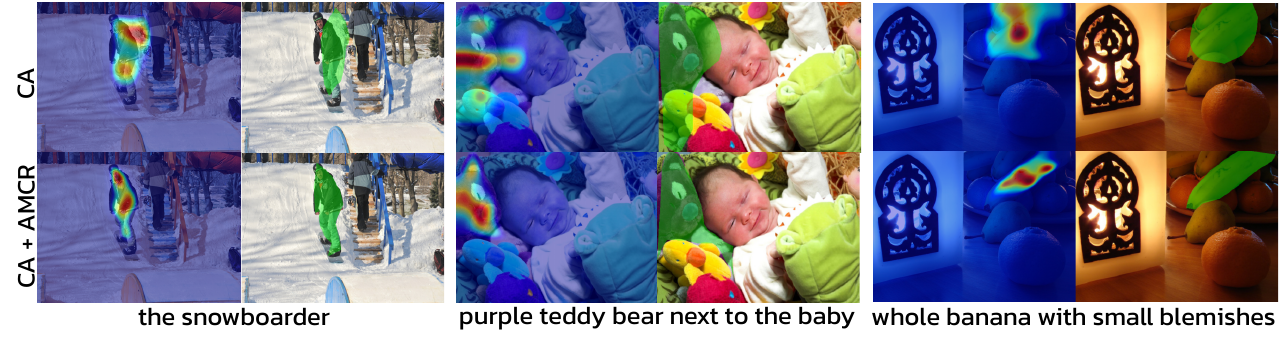}
    \caption{\textbf{Qualitative differences between cross-attention maps and predicted masks in the presence and absence of AMCR}. Without AMCR, some regions outside the object boundary are attended which affects the quality of predicted masks.}
    \label{fig:CA_AMCR_fig}
\end{figure}
% Some existing Vision-Language-Pretrained (VLP) models performing used for visual grounding tasks (e.g., REC, RES) achieve cross-modality fusion through learning stacked fusion layers on top of dual encoders \cite{uniter,ernievl,villa,vilbert}. Although such a scheme results in efficient cross-modal interaction, it introduces a large number of fusion parameters. Considering the above factor, we inject cross-modality fusion through attention into the layers of uni-modal feature extractors. This results in an efficient cross-modality fusion with fewer parameters than the stacked encoder scheme.
\noindent \textbf{3.1.2 Attention Mask Consistency Regularization.}
Feature fusion through cross-attention strategy is highly effective in attending those parts of the image which the texts are referring to. However, in Figure \ref{fig:CA_AMCR_fig}, we observe that the attended regions are scattered across the objects including some background pixels (more pronounced in WSRES task). The lack of fidelity in being able to attend fine-grained information may be attributed to the lack of training data (in WSRES). It has also been found in \cite{raghu2021vision} that vision transformers tend to demonstrate more uniform representations across all layers, enabled by early aggregation of global information and propagation of features from lower to higher layers. Moreover, in the \textit{absence} of abundant masks (and boxes), the localization capability of the model needs to be further enhanced, especially, to yield superior pseudo-masks for the bootstrapping stage. Therefore, to address such issues, we formulate Attention Mask Consistency Regularization (AMCR) Loss to localize fine-grained cross-attention within the target object:
\vspace{-2mm}
\begin{gather}\label{eq:amcr}
\mathcal{L}_{AMCR} = \sum_{b = 1}^{{N}}\underbrace{\left(1-\frac{\sum_{i,j}\mathcal{A}_{i,j}\mathcal{M}_{i,j}}{\sum_{i,j}\mathcal{A}_{i,j}}\right)}_{\text{Localization}}
+ \underbrace{\psi \text{KL}\left(\textbf{U}_{N}||\textbf{Q}_{{N}}\right)}_{\text{Collapse-Reduction}} \\
\textbf{Q}_{{N}} = \left\{\left(\frac{n(\mathcal{A}_{i,j})}{\sum_{i,j}\mathcal{M}_{i,j}}\right)_{b} \forall \; b \in N \right\} \nonumber
\end{gather}
where $\mathcal{A}_{i,j}$ and $\mathcal{M}_{i,j}$ represent the Cross-Attention map from the last layer and Mask, respectively, $(i,j)$ represents pixel location, KL denotes KL divergence loss, $\textbf{U}_{{N}}(0,1)$ denotes uniform distribution of minimum value 0 and maximum value 1, $\textbf{Q}_{{N}}$ represents computed normalized frequency distribution over batch with a size $N$. $\psi$ is the loss-balancing term which we empirically set to 0.001.
% That is, although the attention map is residing within the mask of the object it's size may be small.
$\mathcal{L}_{AMCR}$ consists of two terms: a Localization term and a Collapse-Reduction term. The localization term guarantees accurate localization and alignment of the generated cross-attention map within the mask of the object. The collapse-reduction term prevents collapsing of the cross-attention map within the mask. $\mathcal{L}_{AMCR}$ gets added to the Cross-Entropy (CE) loss and the total loss becomes:
\vspace{-2mm}
\begin{equation}\label{eq:loss_final}
    \mathcal{L}_{total} = \mathcal{L}_{CE} + \lambda \cdot \mathcal{L}_{AMCR}
\end{equation}
where $\lambda$ is responsible for weighting the $\mathcal{L}_{AMCR}$ loss term. We provide more details regarding the formulation of $\mathcal{L}_{AMCR}$ in the Supplementary.
\vspace{-1mm}
\begin{quote}
    \makebox[\linewidth]{%
        \colorbox{LightColor}{%
            \hspace*{0mm} % Adjust left alignment
            \begin{minipage}{\dimexpr\linewidth+10\fboxsep\relax} % Adjust width
                % \fontsize{9pt}{10pt}\selectfont % Font settings
                \textbf{Takeaways:} \textbf{(1)} \FFE's \textit{cross-modal fusion} scheme is an \textit{effective}, \textit{flexible} and \textit{parameter efficient} strategy (when compared to VLP models \cite{dou2022empirical, li2021align, glip} using extra fusion-specific layers) and is particularly beneficial in the context of \textit{weakly-supervised} RES task (Tables \ref{table:sotaonres},\ref{table:davishmdb},\ref{tab:CA_AMCR},\ref{tab:XFACT_vs_ALBEF}). \textbf{(2)} The AMCR component in \FFE\ fosters prediction of high quality masks by spatially constraining the cross-attention map within the target object boundary. This improves \textit{cross-modal region-level alignment} quality, especially in the \textit{limited annotations scenario}. Such gains are not achieved upon using L2 loss (Table \ref{tab:AMCR_vs_L2}).
            \end{minipage}%
        }%
    }
\end{quote}
\vspace{-5mm}

\subsection{Weak-Supervision with $\gamma$-Scheduling.}
% We also consider the validation set to be fully available. It is to be noted that our training data comes from the same distribution.
% We generate \emph{pseudo-labels} from the remaining $(100-x)\%$ image-text pairs (without mask annotations) and retrain the model in a semi-supervised manner with these generated pseudo labels in the next step. However, unlike \cite{qu2023learning}, we \emph{do not} have bounding box annotations for the unlabeled samples. Therefore, we propose a Mask Validity Filtering (MVF) approach (discussed in Section \ref{sec:MVF}). 

It is worth reiterating that unlike Partial-RES \cite{qu2023learning}, we \emph{do not have box annotations for the full dataset}. Therefore, we do not \emph{pretrain} our model on any box-prediction tasks such as Referring Expression Comprehension (REC) unlike Partial-RES. Our goal is to best adapt our architecture to the target RES task given limited mask annotations. Particularly, in a dataset consisting of $\mathcal{N}$ training image-referring expression pairs, we consider mask (and box) annotations for $x\%$ data. These $x\%=\{10\%,20\%,30\%\}$ samples are selected randomly from the original training set. Our overall pipeline (Figure \ref{fig:bootstrap}) is as follows:\\
\noindent $\bullet$ \textbf{Step 1: Initial RES Training.} In this initial step, we train \modelname\ using using Equation \ref{eq:loss_final} with $x\%$ labeled mask data on the RES task. Subsequently, we obtain a trained \emph{pseudo-labeler} with updated model parameters.\\
$\bullet$ \textbf{Step 2: Pseudo-labeling.} We infer masks by running inference on the remaining unlabeled $(100-x)\%$ training samples using model trained from Step 1. Inferred masks are subsequently passed through the proposed Mask Validity Filtering (MVF) (Section \ref{sec:MVF}) to verify the validity of these generated masks in a zero-shot fashion. Contour points are next sampled from the valid masks and added to the corresponding image-text pairs as \emph{pseudo-masks}.\\
$\bullet$ \textbf{Step 3: Retraining with $\gamma$-Scheduling.} We retrain \modelname\ (initialized from previous training) with the updated training dataset containing both the $x\%$ Ground Truth (GT) Masks ($M$) and Pseudo-Masks ($\hat{M}$) and minimize the final loss $\mathcal{L}_{\modelname}$ using a Pseudo-Mask Loss weighting hyperparameter $\gamma$:
\vspace{-1mm}
\begin{align}\label{eq:WSL}
\mathcal{L}_{\modelname} = \underbrace{\mathcal{L}_{total}^{GT}}_{\text{Loss with GT Masks}}
+ \gamma \cdot \underbrace{\mathcal{L}_{total}^{Pseudo}}_{\text{Loss with Pseduo-Masks}}
\end{align}

The role of $\gamma$ is to balance the amount of GT masks against that of pseudo-masks during the retraining step. We introduce a $\gamma$-scheduling strategy to systematically change the value of $\gamma$ over a set of iterations with the starting (minimum) value of $\gamma$ being $\gamma_{0}=0.9$ (found empirically through ablation studies) and maximum value being 1.0. The intuition is to provide more weighting to pseudo-masks as the model becomes more confident of the predictions as the training steps progress. We outline the detailed steps in Algorithm 1 in Supplementary.
\vspace{-1mm}
\begin{quote}
    \makebox[\linewidth]{%
        \colorbox{LightColor}{%
            \hspace*{0mm} % Adjust left alignment
            \begin{minipage}{\dimexpr\linewidth+10\fboxsep\relax} % Adjust width
                % \fontsize{9pt}{10pt}\selectfont % Font settings
                \textbf{Takeaway:} Weak-Supervision with $\gamma$-scheduling pipeline boosts \textit{self-labeling capabilities} in the system whereby \textit{filtered} inferred masks are iteratively re-utilized as \textit{pseudo-labels} in the re-training phase, improving mIoU (Tables \ref{table:sotaonres}-\ref{table:davishmdb}). 
            \end{minipage}%
        }%
    }
\end{quote}
\vspace{-5mm}

% In reality, there may be cases where in the presence of two or more similar objects, CLIP scores the one which is not referred to in the text, i.e., the object with a different spatial location and orientation. 
% , extracted using SpaCy. Noun chunks are composed of entities belonging to the root and the leaf nodes.
% Both noun chunks and their associated relationships can be effectively extracted from the dependency parsing trees where the dependency paths define the relationships. Specifically, we consider the most frequently occurring spatial relationships such as \emph{right/east}, \emph{left/west}, \emph{smaller/tinier/further}, \emph{bigger/larger/closer}, \emph{between}, \emph{within/inside}, \emph{above/north/top}, \emph{below/under/south}, \emph{back/behind}, and \emph{front}. 

\subsection{Mask Validity Filtering with \MVF.}
\label{sec:MVF}

Mask Validity Filtering (MVF) is the process of identifying which predicted (inferred) masks actually localize the object (present in the image) being referred to in the text (referring expression) \cite{qu2023learning}. Partial-RES \cite{qu2023learning} achieves this task by using ground-truth bounding-box annotations. However, as noted in Section~\ref{sec:intro}, our proposed setup considers a true \emph{weakly-supervised} scenario where the number of bounding box and mask annotations are equal. Therefore, to address this gap in the literature, we propose a zero-shot approach to obtain these boxes to be used in the MVF stage. This can also be formulated as a Zero-Shot Referring Expression Comprehension (ZS-REC) task. Our proposed MVF approach is composed of two parts - (i) obtaining bounding boxes using ZS-REC, and (ii) validating the inferred masks using the obtained bounding boxes.\\
\noindent \textbf{3.3.1 ZS-REC with \MVF\ module.} We introduce \textbf{Sp}atially \textbf{A}ware \textbf{R}ed-Box \textbf{C}lip (\MVF), a ZS-REC module, comprising two components: a proposal scoring approach with visual prompts (red-box), and a spatial reasoning component accounting for spatial understanding of proposals in a rule-based manner.\\
\underline{\textit{Proposal scoring with red-box prompting:}} Our ZS framework considers an input image and the referring expression along with $\mathcal{N}_{\mathrm{B}}$ box proposals as generated by a pretrained object detector \cite{FasterRCNN}. ZS-REC task can be conveniently converted to a ZS-retrieval task using an ensemble of image (visual) prompts with a single text as described below. Each of these visual prompts (images) must correspond to a single isolated proposal from the original image generating $\mathcal{N}_{\mathrm{B}}$ such images (visual prompts). Each proposal object region is isolated by employing a Gaussian blur on the background and by adding a red box (red border) of uniform thickness surrounding the object proposal. These image prompts along with the referring expression is passed through CLIP to obtain the CLIP score. Notably, using a red border acts as a \textit{positive visual prompt} highlighting the target area, whereas blurring the background reduces the impact of weakly related information. However, these CLIP scores do not consider the spatial understanding of the objects in the different bounding boxes. Therefore, injecting spatial understanding is important which we discuss next.\\
\noindent{\underline{\textit{Spatial Reasoning component:}}} A major limitation of CLIP for the ZS-REC task is the lack of fine-grained spatial understanding. In Figure \ref{fig:mainfig}, the text "\emph{black dog}" refers to the two black dogs in the picture. CLIP will generate high scores for both these proposal regions since both fall under the class "\emph{black dog}". Furthermore, although addition of visual prompt improves detection performance \cite{redcircle}, it neither incorporates nor enhances spatial understanding capabilities of the model. To mitigate this, we introduce a simple rule-based approach. We divide the referring expression into two constituents - noun chunks and their relations. In the previous example, it refers to the "\emph{black dog}" in the picture which is the subject. The phrases connecting these noun chunks form the relationships (e.g., \emph{right/east}, \emph{left/west}, \emph{smaller/tinier/further}, \emph{bigger/larger/closer}, \emph{between}, \emph{within/inside}, \emph{above/north/top}, \emph{below/under/south}, \emph{back/behind}, and \emph{front}). In the presence of two boxes referring to the \emph{black dog}, the probability of the proposal on the left will be the most (as decided based on the location of the centroid of the box computed using the obtained box coordinates). To provide more clarity on this, we consider the example: "\emph{black dog on the left of a white dog}", relation $R$ between two nouns black dog ($X$) and white dog ($Y$) is "\emph{left}". Coordinates of proposals $X$ and $Y$, and relation $R$ are the inputs of Spatial Reasoning component. If center point of box $X$ is to the "\emph{left}" of that of box $Y$, the output probability $\text{Pr}[R(X,Y)]=1$, otherwise 0. These probabilities are multiplied with the CLIP scores of the respective proposals. Note that in rare occasions where the referring texts in the RefCOCO datasets do not have any spatial relations, we resort to using only CLIP scores for scoring the proposals. More examples, with those of complex relations are given in Supplementary. \\
\noindent \textbf{3.3.2 Validation of Inferred Masks with \MVF.} In the Pseudo-labeling step, \modelname\ predicts contour points which are connected to generate binary masks. We generate a bounding box from the outermost (top-most, bottom-most, right-most, left-most) points of each mask and compute Dice Similarity Coefficient (DSC) \cite{dice} between the generated box and the box obtained using the ZS-REC step using \MVF. We reject the noisy pseudo-masks with DSC value less than $\tau=0.1$ (ablation in Supplementary). Contour points are resampled from the filtered pseudo-masks and added to the training set (Figure \ref{fig:bootstrap}).

\vspace{-2mm}
\begin{quote}
    \makebox[\linewidth]{%
        \colorbox{LightColor}{%
            \hspace*{0mm} % Adjust left alignment
            \begin{minipage}{\dimexpr\linewidth+10\fboxsep\relax} % Adjust width
                % \fontsize{9pt}{10pt}\selectfont % Font settings
                \textbf{Takeaways:} \textbf{(1)} Visual prompting with red border together with the \textit{spatial awareness} component in \MVF\ improves mIoU (as demonstrated in Table \ref{tab:ablation_prompt}). \textbf{(2)} Mask Validity Filtering with \MVF\ enhances the quality of training samples in the retraining stage, which has a positive effect on the model performance (Tables \ref{table:sotaonres}-\ref{table:davishmdb}) for the weakly-supervised RES task.
            \end{minipage}%
        }%
    }
\end{quote}
\vspace{-3mm}

\section{Experiments and Results}
\label{sec:experiments}

\subsection{Datasets, Implementation Details and Metrics}
\noindent\textbf{4.1.1 Dataset.} For \textit{RES}, we conduct our experiments on the three major RES datasets: RefCOCO \cite{yu2016modeling}, RefCOCO+ \cite{yu2016modeling}, and RefCOCOg \cite{nagaraja2016modeling,mao2016generation}. For
\textit{Zero-shot Transfer to Referring Video Object Segmentation} (\textit{ZS-R-VOS}), we conduct zero-shot (ZS) evaluations on two popular R-VOS datasets: RefDAVIS17 \cite{refdavis} and JHMDB-Sentences \cite{jhmdb2}.\\
\noindent\textbf{4.1.2 Implementation.} We use AdamW optimizer \cite{adamw} with a batch size of 128. We use an initial learning rate (LR) of 5e-4 with $(\beta_1, \beta_2) = (0.9, 0.98), \epsilon = 10^{-9}$, and Multi-Step LR Warmup of 5 epochs, decay steps of 75, and a decay ratio of 0.1. We use A100 40GB GPUs for all experiments.\\
% \autoref{sec:supp_dataset} and \autoref{sec:supp_implementation}.
\noindent\textbf{4.1.3 Metrics.}
For \textit{RES}, we report mean Intersection-over-Union (mIoU) values. For \textit{ZS-R-VOS}, we report mean of region similarity ($\mathcal{J}$) and contour accuracy ($\mathcal{F}$) denoted as $\mathcal{J}\&\mathcal{F}$ on RefDAVIS17, and mIoU scores on JHMDB-Sentences. Additional dataset and implementation details are provided in Supplementary.

\begin{table}[t]
\scriptsize
\begin{center}
\resizebox{\columnwidth}{!}
{\begin{tabular}{c c |l c|ccc|ccc|ccc}
\toprule
\multicolumn{2}{c|}{\textbf{Label-Rate}} &
 \multirow{2}{*}{\textbf{Method}} & \multirow{2}{*}{\textbf{Vis. Backbone}} & \multicolumn{3}{c|}{\textbf{RefCOCO}} & \multicolumn{3}{c|}{\textbf{RefCOCO+}} & \multicolumn{3}{c}{\textbf{RefCOCOg}} \\
% \cmidrule{4-6}
% \cmidrule{7-9}
% \cmidrule{10-12}
\textbf{Mask} & \textbf{BBox} & & & \textbf{val} & \textbf{testA} & \textbf{testB} & \textbf{val} & \textbf{testA} & \textbf{testB} & \textbf{val-g} & \textbf{val-u} & \textbf{test-u} \\
\midrule
% DMN~\cite{dmn} & RN101 & 49.78 & 54.83 & 45.13 & 38.88 & 44.22 & 32.29 & 36.76 & - & - \\
\multicolumn{13}{c}{\textit{Fully Supervised Models}} \\
\midrule
% \multirow{15}{*}{100\%} & \multirow{15}{*}{100\%} & CMSA~\cite{cmsa} & RN101 & 58.32 & 60.61 & 55.09 & 43.76 & 47.60 & 37.89 & 39.98 & - & - \\
% & & STEP~\cite{step} & RN101 & 60.04 & 63.46 & 57.97 & 48.19 & 52.33 & 40.41 & 46.40 & - & - \\
% & & CMPC~\cite{cmpc} & DN53 & 61.36 & 64.53 & 59.64 & 49.56 & 53.44 & 43.23 & 49.05 & - & - \\
% & & LSCM~\cite{lscm} & DLA34 & 61.47 & 64.99 & 59.55 & 49.34 & 53.12 & 43.50 & 48.05 & - & - \\
% & & MCN~\cite{mcn} & DN53 & 62.44 & 64.20 & 59.71 & 50.62 & 54.99 & 44.69 & - & 49.22 & 49.40 \\
% & & BUSNet~\cite{busnet} & RN101 & 63.27 & 66.41 & 61.39 & 51.76 & 56.87 & 44.13 & 50.56 & - & - \\
\multirow{9}{*}{100\%} & & LTS~\cite{lts} & DN53 & 65.43 & 67.76 & 63.08 & 54.21 & 58.32 & 48.02 & - & 54.40 & 54.25 \\
& & VLT~\cite{vlt} & DN56 & 65.65 & 68.29 & 62.73 & 55.50 & 59.20 & 49.36 & 49.76 & 52.99 & 56.65 \\
& & ResTR~\cite{kim2022restr} & ViT-B & 67.22 & 69.30 & 64.45 & 55.78 & 60.44 & 48.27 & 54.48 & - & - \\
& & SeqTR~\cite{zhu2022seqtr} & DN53 & 67.26 & 69.79 & 64.12 & 54.14 & 58.93 & 48.19 & - & 55.67 & 55.64 \\
& & \CC{}\textbf{\modelname\ (Ours)} & \CC{}\textbf{Swin-B} & \CC{}\textbf{73.35} & \CC{}\textbf{75.02} & \CC{}\textbf{70.71} & \CC{}\textbf{63.03} & \CC{}\textbf{65.81} & \CC{}\textbf{57.64} & \CC{}- & \CC{}\textbf{62.42} & \CC{}\textbf{62.74} \\
% \cmidrule{3-13}
& & \bf \textcolor{blue}{$\Delta_{\text{Ours - SeqTR}}$} & \textcolor{blue}{-} & \textcolor{blue}{6.09} \textcolor{blue}{$\uparrow$} & \textcolor{blue}{5.23} \textcolor{blue}{$\uparrow$} & \textcolor{blue}{6.59} \textcolor{blue}{$\uparrow$} & \textcolor{blue}{8.89} \textcolor{blue}{$\uparrow$} & \textcolor{blue}{6.88} \textcolor{blue}{$\uparrow$} & \textcolor{blue}{9.45} \textcolor{blue}{$\uparrow$} & \textcolor{blue}{-} & \textcolor{blue}{6.75} \textcolor{blue}{$\uparrow$} & \textcolor{blue}{7.10} \textcolor{blue}{$\uparrow$}\\%[0.5ex]
\cdashline{3-13}%[-1.5ex]
& & PolyFormer~\cite{cmsa}$^{\dag}$ & Swin-B & 75.96 & 77.09 & 73.22 & 70.65 & 74.51 & 64.64 & - & 69.36 & 69.88 \\
& & \CC{}\textbf{\modelname$^{\dag}$ (Ours)} & \CC{}\textbf{Swin-B} & \CC{}\textbf{77.21} & \CC{}\textbf{77.83} & \CC{}\textbf{75.72} & \CC{}\textbf{70.78} & \CC{}\textbf{74.53} & \CC{}\textbf{64.88} & \CC{}- & \CC{}\textbf{70.48} & \CC{}\textbf{71.06} \\
% \cmidrule{3-13}
& & \bf \textcolor{blue}{$\Delta_{\text{Ours - PolyFormer}^{\dag}}$} & \textcolor{blue}{-} & \textcolor{blue}{1.25} \textcolor{blue}{$\uparrow$} & \textcolor{blue}{0.74} \textcolor{blue}{$\uparrow$} & \textcolor{blue}{2.50} \textcolor{blue}{$\uparrow$} & \textcolor{blue}{0.13} \textcolor{blue}{$\uparrow$} & \textcolor{blue}{0.02} \textcolor{blue}{$\uparrow$} & \textcolor{blue}{0.24} \textcolor{blue}{$\uparrow$} & \textcolor{blue}{-} & \textcolor{blue}{1.12} \textcolor{blue}{$\uparrow$} & \textcolor{blue}{1.18} \textcolor{blue}{$\uparrow$}\\
\midrule
\multicolumn{13}{c}{\textit{Weakly Supervised Models}} \\
\midrule
% \multirow{3}{*}{30\%} & 30\% & SeqTR~\cite{zhu2022seqtr} & DN53 & 53.24 & 55.58 & 52.19 & 43.16 & 51.57 & 38.51 & - & 46.82 & 46.95 \\
\multirow{3}{*}{30\%} & \textcolor{OrangeRed}{100\%} & Partial-RES~\cite{qu2023learning}$\spadesuit$ & Swin-B & 66.24 & 68.39 & 63.57 & 54.37 & 58.16 & 47.92 & - & 54.69 & 54.81 \\
& \bf \textcolor{ForestGreen}{30\%} & \CC{}\textbf{\modelname\ (Ours)} & \CC{}\textbf{Swin-B} & \CC{}\textbf{67.04} & \CC{}\textbf{69.17} & \CC{}\textbf{64.23} & \CC{}\textbf{54.98} & \CC{}\textbf{59.31} & \CC{}\textbf{48.26} & \CC{}\textbf{-} & \CC{}\textbf{55.72} & \CC{}\textbf{55.83} \\
& & \bf \textcolor{blue}{$\Delta_{\text{Ours - Partial-RES}}\spadesuit$} & \textcolor{blue}{-} & \textcolor{blue}{0.80} \textcolor{blue}{$\uparrow$} & \textcolor{blue}{0.78} \textcolor{blue}{$\uparrow$} & \textcolor{blue}{0.66} \textcolor{blue}{$\uparrow$} & \textcolor{blue}{0.61} \textcolor{blue}{$\uparrow$} & \textcolor{blue}{1.15} \textcolor{blue}{$\uparrow$} & \textcolor{blue}{0.34} \textcolor{blue}{$\uparrow$} & \textcolor{blue}{-} & \textcolor{blue}{1.03} \textcolor{blue}{$\uparrow$} & \textcolor{blue}{1.02} \textcolor{blue}{$\uparrow$}\\
\midrule
% \multirow{3}{*}{20\%} & 20\% & SeqTR~\cite{zhu2022seqtr} & DN53 & - & - & - & - & - & - & - & - & - \\
\multirow{3}{*}{20\%}& \textcolor{OrangeRed}{100\%} & Partial-RES~\cite{qu2023learning}$\spadesuit$ & Swin-B & 65.20 & 67.43 & 62.85 & 53.78 & 57.52 & 47.39 & - & 53.94 & 54.02 \\
& \bf \textcolor{ForestGreen}{20\%} & \CC{}\textbf{\modelname\ (Ours)} & \CC{}\textbf{Swin-B} & \CC{}\textbf{65.88} & \CC{}\textbf{67.96} & \CC{}\textbf{63.24} & \CC{}\textbf{54.23} & \CC{}\textbf{58.07} & \CC{}\textbf{47.67} & \CC{}\textbf{-} & \CC{}\textbf{54.45} & \CC{}\textbf{54.61} \\
& & \bf \textcolor{blue}{$\Delta_{\text{Ours - Partial-RES}}\spadesuit$} & \textcolor{blue}{-} & \textcolor{blue}{0.68} \textcolor{blue}{$\uparrow$} & \textcolor{blue}{0.53} \textcolor{blue}{$\uparrow$} & \textcolor{blue}{0.39} \textcolor{blue}{$\uparrow$} & \textcolor{blue}{0.45} \textcolor{blue}{$\uparrow$} & \textcolor{blue}{0.55} \textcolor{blue}{$\uparrow$} & \textcolor{blue}{0.28} \textcolor{blue}{$\uparrow$} & \textcolor{blue}{-} & \textcolor{blue}{0.51} \textcolor{blue}{$\uparrow$} & \textcolor{blue}{0.59} \textcolor{blue}{$\uparrow$}\\
\midrule
% \multirow{3}{*}{10\%} & 10\% & SeqTR~\cite{zhu2022seqtr} & DN53 & 33.61 & 37.49 & 32.17 & 30.55 & 34.32 & 23.69 & - & 28.95 & 28.41 \\
\multirow{3}{*}{10\%} & \textcolor{OrangeRed}{100\%} & Partial-RES~\cite{qu2023learning}$\spadesuit$ & Swin-B & 64.01 & 65.89 & 61.68 & 52.85 & 56.01 & 46.27 & - & 52.73 & 52.68 \\
& \bf \textcolor{ForestGreen}{10\%} & \CC{}\textbf{\modelname\ (Ours)} & \CC{}\textbf{Swin-B} & \CC{}\textbf{64.02} & \CC{}\textbf{65.91} & \CC{}\textbf{61.76} & \CC{}\textbf{52.98} & \CC{}\textbf{56.24} & \CC{}\textbf{46.48} & \CC{}\textbf{-} & \CC{}\textbf{52.91} & \CC{}\textbf{52.94} \\
& & \bf \textcolor{blue}{$\Delta_{\text{Ours - Partial-RES}}\spadesuit$} & \textcolor{blue}{-} & \textcolor{blue}{0.01} \textcolor{blue}{$\uparrow$} & \textcolor{blue}{0.02} \textcolor{blue}{$\uparrow$} & \textcolor{blue}{0.08} \textcolor{blue}{$\uparrow$} & \textcolor{blue}{0.13} \textcolor{blue}{$\uparrow$} & \textcolor{blue}{0.23} \textcolor{blue}{$\uparrow$} & \textcolor{blue}{0.21} \textcolor{blue}{$\uparrow$} & \textcolor{blue}{-} & \textcolor{blue}{0.18} \textcolor{blue}{$\uparrow$} & \textcolor{blue}{0.26} \textcolor{blue}{$\uparrow$}\\
\bottomrule
\end{tabular}}
\end{center}
\caption{\textbf{Comparison with the state-of-the-arts on the RES task.} \modelname\ substantially outperforms SOTA SeqTR \cite{zhu2022seqtr} in the fully-supervised benchmark. \modelname\ also yields significant gains over baseline Partial-RES \cite{qu2023learning} even without using 100\% box annotations in the WSRES task. $\dag$ means trained on extra data combining RefCOCO datasets \cite{polyformer}. $\spadesuit$ indicates our reimplementation of Partial-RES with Swin-B backbone where we get better mIoUs than their reported values \cite{qu2023learning}. Differences shown in \textcolor{blue}{blue}.}
\label{table:sotaonres}
\vspace{-4mm}
\end{table}
\subsection{Main Results}
\noindent \textbf{4.2.1 Referring Expression Segmentation (RES).} We compare \modelname\ with the state-of-the-art methods on the RefCOCO/+/g validation and test sets and report the mean IoU (mIoU) in the Table \ref{table:sotaonres} under different label-rates.\\
\underline{\textit{\textbf{Full Supervision}}}: \modelname\ outperforms existing benchmark methods on each split of the three datasets by a significant margin (Table \ref{table:sotaonres}). For a fair comparison with PolyFormer \cite{polyformer}, we additionally train \modelname\ on a combined dataset after removing all data from validation and test sets \cite{polyformer}. Unlike PolyFormer, \modelname\ is not pretrained on the REC task. However, \modelname\ still outperforms PolyFormer on all three datasets. Note that we report fully-supervised results to show an upper-bound on the performance of the weakly-supervised models.\\
\underline{\textit{\textbf{Weak Supervision}}:} We consider label rates of 10\%, 20\%, and 30\% for the WSRES task (Table \ref{table:sotaonres}). Despite methods such as Partial-RES using 100\% box annotations, \modelname\ shows substantial improvements over Partial-RES (note that Partial-RES originally used DN-53 backbone and we re-implemented with Swin-B backbone for fair comparisons) for all three label-rates. With just 30\% labeled data and without any REC pretraining, \modelname\ often surpasses the fully-supervised performance of SeqTR on RefCOCO/+g validation sets (e.g., 59.31 vs 58.93 on RefCOCO+@testA, 55.83 vs 55.64 on RefCOCOg@test sets). Weakly-supervised \modelname\ significantly outperforms other fully-supervised baselines (e.g., VLT \cite{vlt} and LTS \cite{lts}). These results bespeak the role of our proposed \FFE\ (containing cross-modal fusion together with AMCR) and \MVF\ modules in obtaining state-of-the-art results in the limited annotation scenarios.\\
\noindent \textbf{4.2.2 Zero-shot Referring Video Object Segmentation (ZS-R-VOS).} We conduct evaluation on the R-VOS datasets (RefDAVIS17 and JHMDB) as a ZS transfer task. We consider the video frames as a sequence of images without involving any temporal information while predicting masks. As shown in Table \ref{table:davishmdb}, \modelname\ achieves state-of-the-art results with just the spatial information, displaying strong generalization capabilities.

We also conduct evaluations on the \textit{ZS-REC} task with our proposed \MVF\ module which significantly outperforms SOTA RedCircle \cite{redcircle} achieving 39.3\% and 16.2\% gains on RefCOCO@testB and RefCOCO+@testB sets respectively. Results and visualizations are given in Supplementary.
% \textbf{ZS-REC based MVF using \MVF.} MVF stage is responsible for filtering pseudo-masks to be used in the semi-supervised retraining stage. Thus, we evaluate the performance of \MVF module in ZS-REC task on RefCOCO/+/g val and test sets in \autoref{table:zsrec}. We notice significant improvements across all datasets surpassing existing SOTA methods, displaying tremendous generalization capabilities.
% It is even better than the state-of-the-art ReferFormer \cite{referformer}, which is fully trained on R-VOS data.

\begin{table}[!t]
    \centering
    % Left table
    \begin{minipage}{0.52\textwidth}
        \centering
        % \setlength{\tabcolsep}{3pt}
        % \fontsize{10}{9}\selectfont
\resizebox{\columnwidth}{!}
{\begin{tabular}{l c c |c|c}
\toprule
\multirow{2}{*}{\textbf{Method}} & 
\multirow{2}{*}{\textbf{Vis. Backbone}} &
\multirow{2}{*}{\textbf{Eval.}} &
\textbf{RefDAVIS17} &
\textbf{JHMDB}\\
& & & $\mathcal{J}\&\mathcal{F}$ & mIoU\\
% \cmidrule{6-8}
% \cmidrule{9-11}
\midrule
% DMN~\cite{dmn} & RN101 & 49.78 & 54.83 & 45.13 & 38.88 & 44.22 & 32.29 & 36.76 & - & - \\
\multicolumn{5}{c}{\textit{Fully Supervised Models (Label-Rate = 100\%)}} \\
\midrule
% CMSA~\cite{cmsa} & RN50 & FT & 34.7 &    - \\
% CMSA + RNN~\cite{cmsa} & RN50 & FT & 40.2 &  - \\
% \demph{URVOS~\cite{seo2020urvos}} & \demph{RN50} & \demph{FT} & \demph{51.5} & \demph{-} \\
% ReferFormer~\cite{referformer} & RN50 & FT & 58.5 & -\\
\demph{ReferFormer~\cite{referformer}} & \demph{Swin-L} & \demph{FT} & \demph{60.5} & \demph{-} \\
% ACAN~\cite{acan} & I3D & FT & - & 28.9 \\
% CMPC-V~\cite{cmpcv} & I3D & FT & - & 34.2 \\
\demph{MTTR~\cite{mttr}} & \demph{Vid-Swin-T} & \demph{FT} & \demph{-} & \demph{36.6} \\
\demph{ReferFormer~\cite{referformer}} & \demph{Vid-Swin-B} & \demph{FT} & \demph{\underline{61.1}} &  \demph{\underline{43.7}} \\
\midrule
PolyFormer~\cite{polyformer} & Swin-B & ZS & 60.9 & 42.4 \\
SeqTR~\cite{zhu2022seqtr} & DN-53 & ZS & 53.5 & 34.9 \\
% \CC{}\textbf{\modelname-100 (Ours)} & \CC{}Swin-B & \CC{}ZS & \CC{}\textbf{59.6} & \CC{}\textbf{42.1} \\
% \midrule
\CC{}\textbf{\modelname-100 (Ours)} & \CC{}Swin-B & \CC{}ZS & \CC{}\textbf{61.3} & \CC{}\textbf{43.2} \\
% \textcolor{blue}{$\Delta_{\text{Ours - PolyFormer}}$} & \textcolor{blue}{-} & ZS & \textcolor{blue}{0.4} \textcolor{blue}{$\uparrow$} & \textcolor{blue}{0.8}\textcolor{blue}{$\uparrow$}\\
\textcolor{blue}{$\Delta_{\text{Ours - SeqTR}}$} & \textcolor{blue}{-} & ZS & \textcolor{blue}{7.8} \textcolor{blue}{$\uparrow$} & \textcolor{blue}{8.3}\textcolor{blue}{$\uparrow$}\\
\midrule
\multicolumn{5}{c}{\textit{Weakly Supervised Models}} \\
\midrule
Partial-RES-30~\cite{qu2023learning} & Swin-B & ZS & 52.3 & 34.8 \\
\CC{}\textbf{\modelname-30 (Ours)} & \CC{}Swin-B & \CC{}ZS & \CC{}\textbf{55.3} & \CC{}\textbf{38.1} \\
\textcolor{blue}{$\Delta_{\text{Ours - Partial-RES-30}}$} & \textcolor{blue}{-} & ZS & \textcolor{blue}{3.0} \textcolor{blue}{$\uparrow$} & \textcolor{blue}{3.3}\textcolor{blue}{$\uparrow$}\\
\midrule
Partial-RES-10~\cite{qu2023learning} & Swin-B & ZS & 51.5 & 34.3 \\
\CC{}\textbf{\modelname-10 (Ours)} & \CC{}Swin-B & \CC{}ZS & \CC{}\textbf{53.1} & \CC{}\textbf{36.4} \\
\textcolor{blue}{$\Delta_{\text{Ours - Partial-RES-10}}$} & \textcolor{blue}{-} & ZS & \textcolor{blue}{1.6} \textcolor{blue}{$\uparrow$} & \textcolor{blue}{2.1}\textcolor{blue}{$\uparrow$}\\
\bottomrule
\end{tabular}}
\caption{\textbf{Evaluation performance of \modelname\ on RefDAVIS17 and JHMDB validation datasets.} \modelname\ achieves significant gains in zero-shot settings for both datasets. FT and ZS indicate Fine-Tuned and Zero-Shot, respectively. Differences in \textcolor{blue}{blue}.}
\label{table:davishmdb}
    \end{minipage}
\hfill
\begin{minipage}{0.47\textwidth}
        \centering
        % \setlength{\tabcolsep}{0pt}
        % \fontsize{10}{9}\selectfont
        \renewcommand{\arraystretch}{1.21}
\resizebox{\columnwidth}{!}{\begin{tabular}{c | c c | c c c}

\toprule

\multirow{2}{2.0 cm}{\bf \centering Label-Rate} & \multirow{2}{2.0 cm}{\bf \centering Cross-Attn.} & \multirow{2}{1.5 cm}{\bf \centering AMCR} & \multicolumn{3}{c}{\bf \centering RefCOCO} \\

& & & \textbf{val} & \textbf{testA } & \textbf{testB} \\

\midrule 

\multirow{3}{*}{100} & \textcolor{OrangeRed}{\ding{55}} & \textcolor{OrangeRed}{\ding{55}} & 68.84 & 70.59 & 65.85\\
 & \textcolor{ForestGreen}{\ding{51}} & \textcolor{OrangeRed}{\ding{55}} & 72.41 & 74.65 & 69.96 \\
 & \CC{}\textcolor{ForestGreen}{\ding{51}} & \CC{}\textcolor{ForestGreen}{\ding{51}} & \CC{}\textbf{73.35} & \CC{}\textbf{75.02} & \CC{}\textbf{70.71}\\
 \midrule
 \multirow{3}{*}{30} & \textcolor{OrangeRed}{\ding{55}} & \textcolor{OrangeRed}{\ding{55}} & 60.91 & 64.06 & 57.73\\
 & \textcolor{ForestGreen}{\ding{51}} & \textcolor{OrangeRed}{\ding{55}} & 64.72 & 67.55 & 60.91 \\
 & \CC{}\textcolor{ForestGreen}{\ding{51}} & \CC{}\textcolor{ForestGreen}{\ding{51}} & \CC{}\textbf{67.04} & \CC{}\textbf{69.17} & \CC{}\textbf{64.23} \\
\midrule
 \multirow{3}{*}{20} & \textcolor{OrangeRed}{\ding{55}} & \textcolor{OrangeRed}{\ding{55}} & 58.84 & 61.79 & 57.02\\
 & \textcolor{ForestGreen}{\ding{51}} & \textcolor{OrangeRed}{\ding{55}} & 61.25 & 64.68 & 59.47 \\
 & \CC{}\textcolor{ForestGreen}{\ding{51}} & \CC{}\textcolor{ForestGreen}{\ding{51}} & \CC{}\textbf{65.88} & \CC{}\textbf{ 67.96} & \CC{}\textbf{63.24} \\
\midrule
\multirow{3}{*}{10} & \textcolor{OrangeRed}{\ding{55}} & \textcolor{OrangeRed}{\ding{55}} & 54.06 & 55.64 & 52.61\\
 & \textcolor{ForestGreen}{\ding{51}} & \textcolor{OrangeRed}{\ding{55}} & 60.11 & 62.18 & 57.95 \\
 & \CC{}\textcolor{ForestGreen}{\ding{51}} & \CC{}\textcolor{ForestGreen}{\ding{51}} & \CC{}\textbf{64.02} & \CC{}\textbf{65.91} & \CC{}\textbf{61.76} \\

\bottomrule
\end{tabular}}
\caption{\textbf{Impact of Cross-Attention and AMCR (in the \FFE\ module) on \modelname's performance} evaluated on RefCOCO val, testA, and testB sets for 100\%, 30\%, 20\%, and 10\% label-rates. Best results are highlighted.}
\label{tab:CA_AMCR}
\end{minipage}
\vspace{-4mm}
\end{table}
% \end{table}

\subsection{Ablation Study}
\label{sec:ablations}
\noindent \textbf{4.3.1 Impact of \FFE.} We show the impact of \FFE\ containing Cross-Attention (CA) based fusion and the AMCR components in Table \ref{tab:CA_AMCR}. Inclusion of CA is shown to improve mIoU values consistently across varying label-rates. The impact of AMCR is more pronounced in the cases of limited annotations. For instance, with a label-rate of 100\%, addition of AMCR boosts mIoU by 0.94 points on RefCOCO@val, whereas with a label-rate of 10\%, this difference increases substantially to 3.91. Additionally, in Figure \ref{fig:CA_AMCR_fig} we demonstrate that incorporating AMCR qualitatively improves both the cross-attention maps and the predicted masks, highlighting the efficacy of AMCR in our pipeline. We further assess the importance of AMCR loss balancing factor ($\lambda$) in the Figure \ref{fig:lambda} when evaluated on RefCOCO@val. Increasing $\lambda$ improves mIoU initially and the maximum is achieved at 0.4, beyond which the performance drops significantly. Moreover, in Table \ref{tab:XFACT_vs_ALBEF} we show that despite being lightweight, our proposed fusion scheme in \FFE\ is superior in performance as compared to (conventional) ALBEF-like fusion which contains extra fusion-specific layers.\\
\begin{figure}[!t]
  \begin{minipage}[b]{0.53\linewidth}
    \centering
    % \includegraphics[width=0.9\linewidth]{Figures/Ablations_plot.png}
    % \vspace{-5mm}
	\centering
    \hspace{-1.0em}
	\begin{subfigure}[b]{0.53\linewidth}
		\resizebox{1.0\linewidth}{!}{
			\begin{tikzpicture}
	\begin{axis} [
		axis x line*=bottom,
		axis y line*=left,
		legend pos=north east,
		ymin=10, ymax=70,
		xmin=0, xmax=7,
		xticklabel={\pgfmathparse{\tick}\pgfmathprintnumber{\pgfmathresult}},
		xtick={1,3,5,7},
		ytick={35,45,55,65},
        xlabel={\footnotesize{run}},
        ylabel={\footnotesize{mIoU}},
        xlabel shift=-4pt,
        ylabel shift=-5pt,
		width=\linewidth,
		legend style={cells={align=left}},
		label style={font=\scriptsize},
		tick label style={font=\scriptsize},
		legend style={draw=none,at={(0.35,0.35)},anchor=west},
		]

        \addplot[mark=*,mark options={scale=0.5, fill=Dark},style={thick},Dark] plot coordinates {
			(1, 56.01)
            (2, 64.71)
			(3, 66.65)
			(4, 67.04)
		};
        \addlegendentry{\scriptsize{$\gamma_{0}=0.9$}}

        \addplot[mark=*,mark options={scale=0.5, fill=Salmon},style={thick},Salmon] plot coordinates {
			(1, 56.01)
            (2, 63.24)
			(3, 64.78)
			(4, 65.31)
		};

        \addlegendentry{\scriptsize{$\gamma_{0}=0.8$}}

		\addplot[mark=*,mark options={scale=0.5, fill=Dark},style={dashed},Dark] plot coordinates {
			(1, 37.84)
            (2, 48.02)
            (3, 55.93)
            (4, 58.78)
			(5, 61.09)
			(6, 63.67)
            (7, 64.02)
		};  

        \addplot[mark=*,mark options={scale=0.5, fill=Salmon},style={dashed},Salmon] plot coordinates {
			(1, 37.84)
            (2, 44.57)
            (3, 51.64)
            (4, 55.61)
			(5, 59.53)
			(6, 61.68)
            (7, 62.10)
		};

		\label{pgf:ek100_map}
	\end{axis}
\end{tikzpicture}%
		}
		\caption{mIoU vs runs.}
        \label{fig:runs}
	\end{subfigure}
 \hspace{-1.5em}
	\begin{subfigure}[b]{0.53\linewidth}
		\resizebox{1.0\textwidth}{!}{
			\begin{tikzpicture}
	\begin{axis} [
		axis x line*=bottom,
		axis y line*=left,
		legend pos=north east,
		ymin=30, ymax=70,
		xmin=0, xmax=0.6,
		xticklabel={\pgfmathparse{\tick}\pgfmathprintnumber{\pgfmathresult}},
		xtick={0.2,0.4,0.6},
		ytick={35,45,55,65},
        xlabel={\footnotesize{$\lambda$}},
        ylabel={\footnotesize{mIoU}},
        xlabel shift=-5pt,
        ylabel shift=-5pt,
		width=\linewidth,
		legend style={cells={align=left}},
		label style={font=\scriptsize},
		tick label style={font=\scriptsize},
		legend style={draw=none,at={(0.30,0.35)},anchor=west},
		]

        \addplot[mark=*,mark options={scale=0.5, fill=Dark},style={thick},Dark] plot coordinates {
			(0.0, 64.72)
            (0.2, 66.45)
			(0.4, 67.04)
			(0.6, 64.97)
		};
        \addlegendentry{\scriptsize{\modelname-30}}

        \addplot[mark=*,mark options={scale=0.5, fill=Dark},style={dashed},Dark] plot coordinates {
			(0.0, 60.11)
            (0.2, 62.85)
			(0.4, 64.02)
			(0.6, 61.34)
		};
        \addlegendentry{\scriptsize{\modelname-10}}

		% \addplot[dashed,gray] plot coordinates {
		% 	(0, 26.0)
		% 	(100, 26.0)
		% };
		% \addplot[mark=asterisk,color_ao,mark options={scale=2,thick},only marks] plot coordinates {
		% 	(1, 67.26)
		% };
		\label{pgf:ek100_map}
	\end{axis}
\end{tikzpicture}%
		}
		\caption{mIoU vs $\lambda$.}
        \label{fig:lambda}
	\end{subfigure}
    % \vspace{-3mm}
    \vspace{-3.0mm}
    \captionof{figure}{\textbf{Impact of $\gamma$-scheduling under different initial values ($\gamma_{0}$) and AMCR balancing factor ($\lambda$)} when evaluated on RefCOCO@val at 30\% and 10\% mask labels.}
  \end{minipage}\hfill
\begin{minipage}[b]{0.45\linewidth}
    \centering
    % \vspace{-15mm}
    \renewcommand{\arraystretch}{0.90}
\resizebox{\columnwidth}{!}{\begin{tabular}{@{} c c c c | c c c @{}}

\toprule

\textbf{Red-box} & \textbf{Spatial} & \textbf{Blur} & \textbf{Crop} & \textbf{Ref@val} & \textbf{Ref+@val} & \textbf{Refg@val} \\

\midrule

\textcolor{OrangeRed}{\ding{55}} & \textcolor{OrangeRed}{\ding{55}} & \textcolor{OrangeRed}{\ding{55}} & \textcolor{ForestGreen}{\ding{51}} & 63.11 & 47.48 & 49.43\\
\textcolor{OrangeRed}{\ding{55}} & \textcolor{OrangeRed}{\ding{55}} & \textcolor{ForestGreen}{\ding{51}} & \textcolor{OrangeRed}{\ding{55}} & 63.78 & 49.66 & 51.48\\
\textcolor{ForestGreen}{\ding{51}} & \textcolor{OrangeRed}{\ding{55}} & \textcolor{OrangeRed}{\ding{55}} & \textcolor{OrangeRed}{\ding{55}} & 64.35 & 51.16 & 52.87\\
\textcolor{ForestGreen}{\ding{51}} & \textcolor{OrangeRed}{\ding{55}} & \textcolor{ForestGreen}{\ding{51}} & \textcolor{OrangeRed}{\ding{55}} & 65.29 & 52.51 & 54.22\\
\textcolor{OrangeRed}{\ding{55}} & \textcolor{ForestGreen}{\ding{51}} & \textcolor{OrangeRed}{\ding{55}} & \textcolor{ForestGreen}{\ding{51}} & 64.12 & 50.19 & 51.80\\
\textcolor{OrangeRed}{\ding{55}} & \textcolor{ForestGreen}{\ding{51}} & \textcolor{ForestGreen}{\ding{51}} & \textcolor{OrangeRed}{\ding{55}} & 65.94 & 52.83 & 54.13\\
\rowcolor{Light}
\textcolor{ForestGreen}{\ding{51}} & \textcolor{ForestGreen}{\ding{51}} & \textcolor{ForestGreen}{\ding{51}} & \textcolor{OrangeRed}{\ding{55}} & \textbf{67.04} & \textbf{54.98} & \textbf{55.72}\\

\bottomrule
\end{tabular}}
    \vspace{-3.0mm}
    \captionof{table}{\textbf{mIoUs with various prompts for \MVF} at 30\% label-rate. Best results are obtained when both red-box and blurring are used with the spatial reasoning component.}
\label{tab:ablation_prompt}
  \end{minipage}
\end{figure}
\begin{table}[!t]
    \centering
    % Left table
    \begin{minipage}{0.54\textwidth}
        \centering
        % \setlength{\tabcolsep}{3pt}
        % \fontsize{10}{9}\selectfont
    \resizebox{\linewidth}{!}{\begin{tabular}{c | c c | c c c}
\toprule
\textbf{Label-Rate} & \textbf{L2} & \textbf{AMCR} & \textbf{Ref@val} & \textbf{Ref+@val} & \textbf{Refg@val} \\
\midrule
\multirow{2}{*}{30} & \textcolor{ForestGreen}{\ding{51}} & $-$ & 63.48 & 51.31 & 52.16\\
& \CC{}$-$ & \CC{}\textbf{\textcolor{ForestGreen}{\ding{51}}} & \CC{}\textbf{67.04} & \CC{}\textbf{54.98} & \CC{}\textbf{55.72}\\
\midrule
\multirow{2}{*}{10} & \textcolor{ForestGreen}{\ding{51}} & $-$ & 59.23 & 48.29 & 48.83\\
& \CC{}$-$ & \CC{}\textbf{\textcolor{ForestGreen}{\ding{51}}} & \CC{}\textbf{64.02} & \CC{}\textbf{52.98} & \CC{}\textbf{52.91}\\
\bottomrule
\end{tabular}}
% \vspace{-3.7mm}
\caption{\textbf{Comparison of proposed AMCR with L2 loss} for 10\% and 30\% annotations.}
% \vspace{-8mm}
\label{tab:AMCR_vs_L2}
    \end{minipage}
\hfill
\begin{minipage}{0.45\textwidth}
        \centering
        % \setlength{\tabcolsep}{0pt}
        % \fontsize{10}{9}\selectfont
        \renewcommand{\arraystretch}{0.8}
\resizebox{\linewidth}{!}{\begin{tabular}{c | c c | c }
\toprule
\textbf{Fusion} & \textbf{Label-} & \textbf{RefCOCOg} & \textbf{$\Delta$ parameter} \\
\textbf{Scheme} & \textbf{Rate} & \textbf{val} & \textbf{increase (\%) $\downarrow$} \\
\midrule
\textbf{\FFE} & \CC{}30 & \CC{}\textbf{55.72} & \multirow{2}{*}{\textbf{1.82}}\\
\textbf{(Ours)} & \CC{}100 & \CC{}\textbf{62.42} & \\
\midrule
\textbf{ALBEF-} & 30 & 54.29 & \multirow{2}{*}{29.65}\\
\textbf{like} & 100 & 60.17 & \\
\bottomrule
\end{tabular}}
% \vspace{-3.7mm}
\caption{\textbf{\FFE\ vs ALBEF-like fusion} for 30\% and 100\% annotations.}
% \vspace{-8mm}
\label{tab:XFACT_vs_ALBEF}
\end{minipage}
\vspace{-8mm}
\end{table}
\hspace{-0.9mm}\noindent\textbf{4.3.2 Retraining with MVF using \MVF.} In the presence of limited mask (and box) annotations, retraining with $\gamma$ scheduling leads to a marked improvement in the mIoU values as depicted in Figure \ref{fig:runs}. Here, we consider label-rates of 30\% (solid lines) and 10\% (dashed lines) with two different $\gamma_{0}$ values denoting the starting values of the $\gamma$ variable. We note that as the number of retraining steps (runs) increases, the mIoU values (for both $\gamma_{0}$ values) in each label-rate increases. However, with $\gamma_{0}=0.9$ we achieve slightly better performance. Additionally, we evaluate the effectiveness of the MVF stage with \MVF\ in Supplementary. Results indicate that without MVF, generated pseudo-labels cannot be validated which negatively impacts model's performance. In contrast, when pseudo-labels are validated using MVF, the mIoU values on RefCOCO@val set are shown to improve by 4.36 mIoU, demonstrating the importance of MVF.\\
\noindent \textbf{4.3.3 Impact of various components of \MVF.} In Table \ref{tab:ablation_prompt} we assess the impact of individual components of \MVF. We note that red-box prompting and blurring together boost performance. Adding spatial awareness via the rule-based reasoning module also substantially improves mIoU on the WSRES task.\\
\noindent \textbf{4.3.4 AMCR vs L2 loss in \FFE.} To evaluate the impact of the AMCR component in \FFE, we analyze \modelname's performance after replacing the proposed formulation of AMCR with a L2 loss. As shown in Table \ref{tab:AMCR_vs_L2}, we observe that using L2 (in place of AMCR) leads to substantially poor performance, as compared to when AMCR is used in \FFE. This corroborates the strong localization capabilities introduced in the system by the AMCR formulation which is not achieved with a L2 loss.\\
\noindent \textbf{4.3.5 Impact of Label Rates.} We notice that as the label-rate increases from 10\% to 100\%, the performance of the model improves significantly across all the datasets (see Table \ref{table:sotaonres} and Figure \ref{fig:teaser}). Remarkably, with just 30\% annotated data, \modelname\ achieves 59.31 mIoU versus 58.93 mIoU obtained by the fully-supervised SeqTR on RefCOCO+@testA set.

We provide additional ablations on the impacts of MVF, gated cross-attention parameter, and DSC thresholding parameter in Supplementary.
% \textbf{\modelname\ vs Partial-RES with Swin Backbone:}\\
% With just 30\% mask-annotated data, \modelname\
% achieves 59.31 mIoU vs 58.93 mIoUs by the fully-supervised SeqTR on RefCOCO+@testA set.

\begin{figure}[!t]
    \centering
    % Left table
    \begin{minipage}{0.52\textwidth}
        \centering
        % \setlength{\tabcolsep}{3pt}
        % \fontsize{10}{9}\selectfont
    \includegraphics[width=\textwidth]{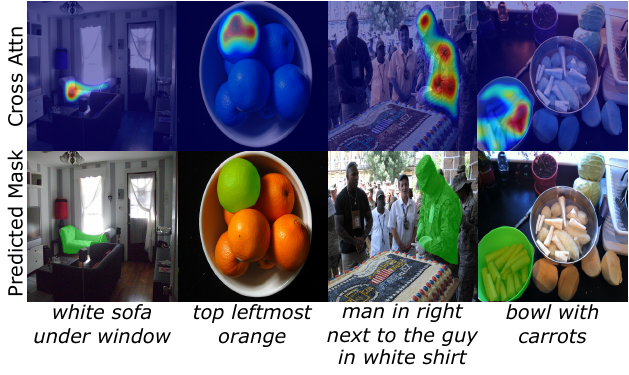}
    \caption{\textbf{Cross-attention Maps and corresponding predictions} showing strong cross-modal alignment learned by \modelname.}
    \label{fig:CA_masks_refcoco}
    \end{minipage}
\hfill
\begin{minipage}{0.47\textwidth}
        \centering
        % \setlength{\tabcolsep}{0pt}
        % \fontsize{10}{9}\selectfont
    \includegraphics[width=\textwidth]{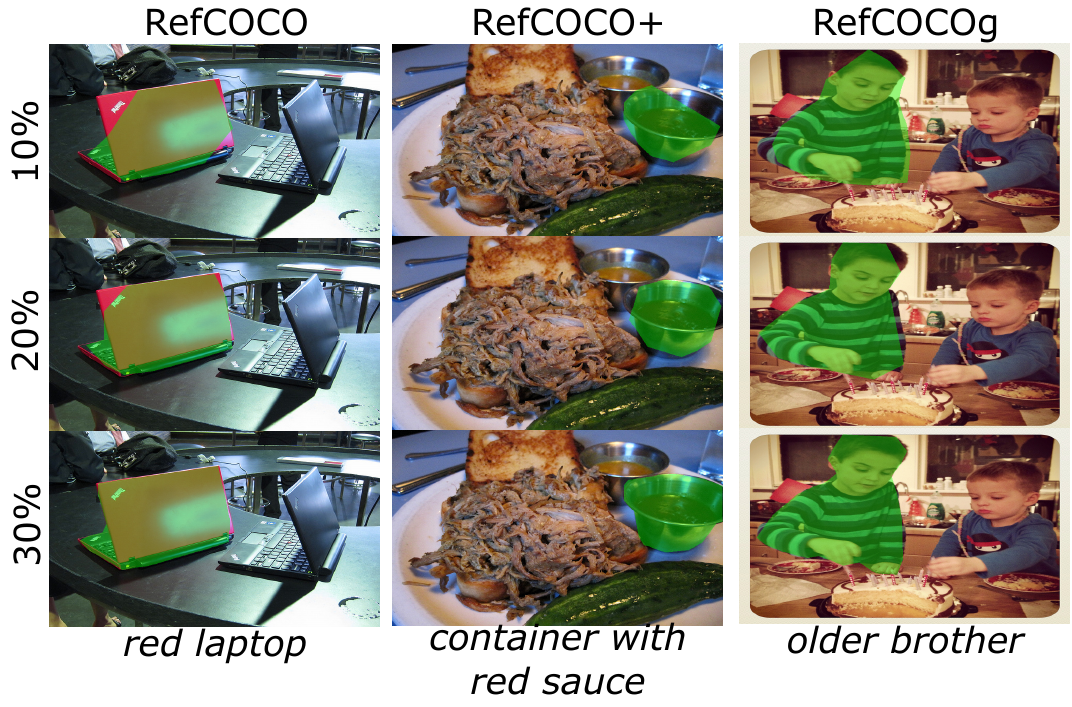}
    \caption{\textbf{Predictions with varying label-rates.} With increasing mask annotations \%, prediction quality improves.}
    \label{fig:masks_label_rates}
\end{minipage}
% \vspace{-5mm}
\end{figure}
% \end{table}
% \vspace{-5mm}
\subsection{Qualitative Assessment and Error Analysis}
\label{sec:qual_analysis}
\begin{figure*}[t]
    \centering
    \includegraphics[width=\textwidth]{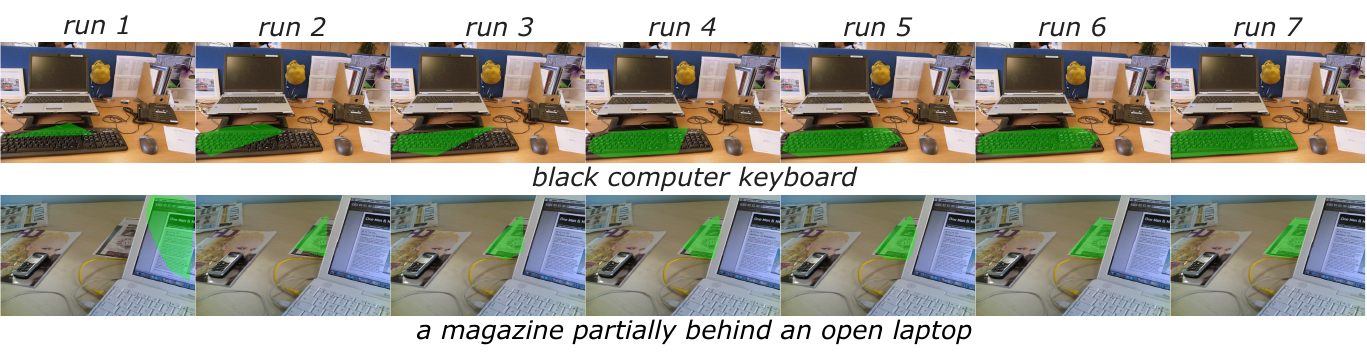}
    \caption{\textbf{Examples of masks with increasing WSRES bootstrapping runs (steps) for 10\% annotations.} We see significant improvements in localization capabilities with an increase in retraining steps illustrating the efficacy our approach.}
    \label{fig:masks_gamma_scheduling_runs}
\end{figure*}

\begin{figure*}[t]
    \centering
    \includegraphics[width=\textwidth]{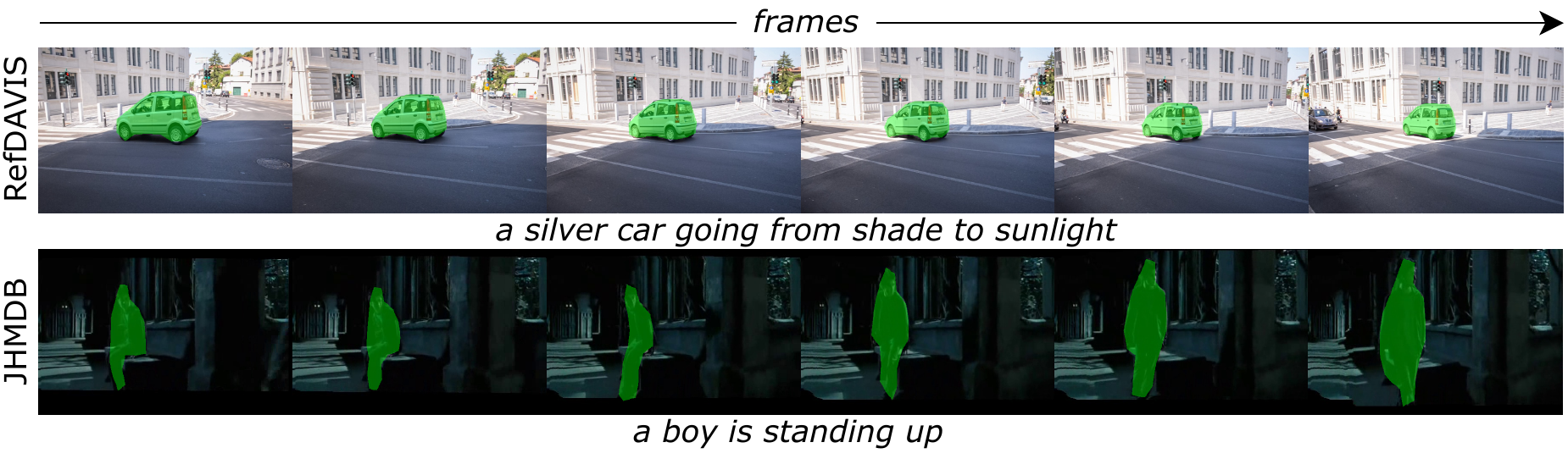}
    \caption{\textbf{Predicted masks on RefDAVIS17 and JHMDB datasets} in a zero-shot setting with \modelname-30 (trained with 30\% annotations).}
    \label{fig:masks_refdavis_jhmdb}
    % \vspace{-4mm}
\end{figure*}

\noindent \textbf{4.4.1 Backbone Cross-Attentions and Predicted Masks.} In Figure \ref{fig:CA_masks_refcoco}, we show how cross-modal attention maps attend to different objects in the images, guided by the referring texts. The cross-attention scores between the image and the associated expression are extracted and bilinearly interpolated to match the image dimension and superimposed on the original image. Notably, strong cross-modal interactions aid in predicting high-quality masks as displayed in Figure \ref{fig:CA_masks_refcoco}. In Figure \ref{fig:masks_refdavis_jhmdb}, we also provide high-quality segmentation masks generated by \modelname\ for the ZS-R-VOS tasks without any finetuning on the respective datasets.\\
\noindent \textbf{4.4.2 Varying Label-Rates.} In Figure \ref{fig:masks_label_rates}, we display segmentation masks under varying label-rates using \modelname. We clearly notice that the quality of masks improve substantially with an increase in the ground-truth annotations.\\
\noindent \textbf{4.4.3 Varying Retraining Steps.} In Figure \ref{fig:masks_gamma_scheduling_runs} we show that with increasing retraining steps (runs), \modelname\ becomes more confident which results in significant qualitative improvements in the predictions. For example, in Figure \ref{fig:masks_gamma_scheduling_runs} although \modelname\ could not recognize \textit{the magazine partially behind laptop} in the first step, it was successful in predicting the mask accurately in the final step - this shows the efficacy of the retraining stage.\\
\noindent \textbf{4.4.4 Comparisons with Partial-RES.} We provide qualitative examples for 30\% label-rate in Figure \ref{fig:teaser}. It is evident from Figure \ref{fig:teaser} that \modelname\ succeeds in challenging cases requiring extensive linguistic understanding (attained with \FFE\ and \MVF\ modules) where Partial-RES \cite{qu2023learning} fails.

We provide extended versions of these qualitative visualizations in Supplementary. Additionally, we show examples of cases in Supplementary where \modelname\ occasionally fails to focus on the referred objects. These cases typically consists of vastly hindered objects in cluttered environments, especially in low-light blurry conditions and tiny objects.

\section{Conclusion}
\label{sec:conclusion}

We present a weakly-supervised learning framework for RES considering limited mask (and box) annotations and employing a contour-based sequence prediction approach. Unlike Partial-RES \cite{qu2023learning}, we do not consider box annotations to be fully available and therefore we do not pretrain our model on the fully-supervised REC task. We incorporate lightweight gated cross-modal attention in the feature backbones along with an Attention Mask Consistency Regularization module to facilitate strong cross-modal alignment and improve the quality of predicted masks. We introduce a bootstrapping pipeline with self-labeling capabilities where pseudo-labels are validated using our proposed Mask Validity Filtering approach. We conduct extensive experiments to demonstrate that \modelname\ consistently achieves state-of-the-art performance across all RES datasets. Finally, we demonstrate excellent generalization capabilities of \modelname\ on \textit{zero-shot} referring video-object segmentation task. Extending our approach to multi-image and video settings can be looked upon as a promising future work.

% Our work being one of the first approaches to tackle weakly-supervised RES task, opens up avenues for further studies in this space. 

% \begin{figure}[!ht]
%      \centering
%      \begin{subfigure}[b]{0.235\textwidth}
%          \centering
%          \includegraphics[width=\textwidth]{Figures/WSRES_and_Partial-RES_1.png}
%          \caption{\textcolor{black}{Samples where both Partial-RES and \modelname\ succeeds.}}
%          \label{fig:}
%      \end{subfigure}
%      % \hfill
%      \begin{subfigure}[b]{0.235\textwidth}
%          \centering
%          \includegraphics[width=\textwidth]{Figures/WSRES_and_Partial-RES_2.png}
%          \caption{\textcolor{black}{Samples where Partial-RES fails, but \modelname\ succeeds.}}
%      \end{subfigure}
%     % \vspace{-.5mm}
%          \caption{}
%     % \vspace{-1.5mm}
%     \label{fig:partial_res_comparison}
% \end{figure}

\bibliographystyle{splncs04}
\bibliography{main}

\end{document}